\documentclass[1p]{elsarticle}

\usepackage{lineno}
\usepackage{hyperref}
\modulolinenumbers[2]

\journal{Journal of \LaTeX\ Templates}
\usepackage{subcaption}
\usepackage{xcolor}
\usepackage{amsmath}
\usepackage{color, soul}
\usepackage{algorithm}
\usepackage{algorithmicx}
\usepackage{algpseudocode}
\usepackage{slashbox}
\usepackage{graphicx}
\usepackage{multirow}
\usepackage{comment}
\usepackage{amsmath}
\usepackage{amssymb}
\usepackage{cleveref}

\begin{document}

\begin{frontmatter}

\title{Integration of Regularized $l$1 Tracking and Instance Segmentation for Video Object Tracking}

\author{Filiz Gurkan  \ \  \ Bilge Gunsel }
\address{Multimedia Signal Proc. and Pattern Recognition Group \\\ Istanbul Technical University, Turkey \\\{gurkanf, gunselb\}@itu.edu.tr}

\begin{abstract}

We introduce a tracking-by-detection method that integrates a deep object detector with a particle filter tracker under the regularization framework where the tracked object is represented by a sparse dictionary. A novel observation model which establishes consensus between the detector and tracker is formulated that enables us to update the dictionary with the guidance of the deep  detector. This yields an efficient representation of the object appearance through the video sequence hence improves robustness to occlusion and pose changes. Moreover we propose a new state vector consisting of translation, rotation, scaling and shearing parameters that allows tracking the deformed object bounding boxes hence  significantly increases robustness to scale changes.
Numerical results reported on challenging VOT2016 and VOT2018 benchmarking data sets demonstrate that the introduced tracker, L1DPF-M, achieves comparable robustness on both data sets while it outperforms state-of-the-art trackers on both data sets where the improvement achieved in success rate at IoU-th=0.5 is  11\% and 9\%, respectively.

\end{abstract}

\begin{keyword}
\texttt{Object tracking, regularized particle filtering, deep object detector, sparse representation.}
\end{keyword}

\end{frontmatter}

\section{Introduction}
\label{intro}
The goal of the video object tracking is to estimate the location of the object in subsequent frames using a target object bounding box (BB) initialized in the first frame of the tracking. Although several methods have been proposed for the video and visual object tracking over the past years, it is still a challenging problem to improve robustness to abrupt appearance changes arising from high motion, occlusion, intensity and scale changes \cite{survey2,survey,sv5,sv4,rev}. 
Tracking methods can be categorized as generative  \cite{mtt,ltpc,low, VRCPF,reversesparse}, discriminative \cite{STR,mil} or combination of both \cite{gen,mcpf,rw, GURKAN}. Under the generative framework, the target appearance is typically represented by a statistical model referred as observation model, while the discriminative framework aims to design a classifier to extract the target from surrounding background. In particular, the observation model aims to reflect the similarity between the initially specified object of interest and the tracked object through the video sequence. The similarity could be modeled either by a color similarity based distance metric \cite{VRCPF} or in terms of a reconstruction error minimized by using a sparse representation of the target object \cite{L1APG,spr,sub,sparss}. As a discriminative detector the statistical classifiers including support vector machines (SVM), online or offline learning schemes and CNN based detectors are widely used  \cite{STR,GURKAN,occ,erdem,Mgnet,CC,im,er}.
Latest work on object tracking focuses on tracking-by-detection (TBD) methods that integrate the discriminative and the generative framework to take advantages of both \cite{GURKAN,TBD1,mdnet,dt,gated,tbd}.

In this paper, we propose a novel TBD method, L1DPF-M, that employs {\em l}1 regularization under Bayesian filtering framework as the generative model while a deep object detector serves as the discriminative model.
We have three main contributions that improve robustness to abrupt appearance changes, scale changes and occlusion. First we define a new observation model that enables us to integrate the object detector and tracker via a maximum likelihood estimator. This enhances localization accuracy of the estimated target object bounding box. Second we formulate a new state vector to tackle the deformed object bounding boxes under affine motion. It is demonstrated that the introduced motion based tracking significantly improves robustness to scale changes and transformations. Moreover, in order to make the tracker robust to high appearance changes, we introduce a target model update scheme that encourages the consensus between the tracker and detector.
Performance evaluation on commonly used Visual Object Tracking benchmark (VOT) 2016 and 2018 data sets \cite{VOT2016,VOT2018} demonstrates that because of the proposed state vector, L1DPF-M outperforms the state-of-the-art trackers especially for the tracking of deformed objects and the introduced fusion scheme provides high localization accuracy while the dictionary update increases robustness to appearance changes. 

Rest of the paper is organized as follows. We summarize the related work in Section \ref{related work} and give the theoretical background in Section \ref{theoretical}. Section \ref{proposed} and \ref{proposed2} formulate the proposed tracker. Section \ref{perf} reports the tracking performance of the proposed method compared to the state-of-the-art methods. Section \ref{conc} summarizes conclusions.

\section{Related Work}
\label{related work}
Particle filters (PF) or sequential Monte Carlo methods (SMC) are one of the commonly used generative methods in tracking because of their effectiveness in representing the object motion with a simple state transition model.
Since they have been first employed for tracking \cite{firsPF}, they are extended in several works \cite{ VRCPF,godsill}.    
Some of the latest TBD methods integrate particle filtering with discriminative techniques including SVM \cite{DSP2} and  discriminative correlation filters \cite{mcpf,hybrid} to improve the tracking accuracy. In \cite{hybrid} the kernel correlation filtering is used to locate object position based on HOG and colour features where the particle filtering is employed to refine the estimated position. Although the modified position and scale information are given as feedback to the correlation filter for training on the target, the tracker suffers from low resolution.
With  the development of deep convolutional neural networks (CNN), a number of methods that integrate a tracking scheme with a deep object detector are developed \cite{GURKAN,im,deepas,Filiz}.  MCPF proposed in \cite{mcpf} performs particle filter tracking guided by a number of multi-task correlation filters that learn dependencies among different features, i.e., HOG features, color-based features and the features extracted by a CNN. MCPF is robust to attributes including background clutter, illumination changes, and rotation but suffers from  motion blur and scale changes. \cite{deepas} concatenates the deep features and the color histograms used by the conventional color particle filtering \cite{cpf} to improve the tracking capability. Differ from the conventional PF where each particle is drawn from a zero mean constant variance Gaussian distribution, \cite{deepas} adaptively changes the mean and variance according to the latest state estimations hence updates the state transition model.  Moreover the observation model of PF is updated by replacing the existing weights with the average of the latest weights to tackle the object appearance changes. 
Finally a rectangle object BB is estimated as a weighted average of the BBs pointed by the particles.
Results are reported on a small data set including 20 videos of OTB-100 data set \cite{otb100}. 
\cite{Filiz} proposes a TBD method that interleaves the variable rate color particle filter and Faster R-CNN deep detector \cite{fastrcnn} based on a decision fusion scheme to dynamically model the target object appearance changes arising from high motion and occlusion. However, in case of the deep detector or PF fails, the interleaving mechanism does not properly work. In order to select the qualified object proposals,   Region Proposal Alignment (RPA) scheme of IDPF-RP introduced in  \cite{GURKAN} applies non-maxima suppression on the proposals generated by Mask R-CNN and VRCPF \cite{VRCPF} that highly improves the localization accuracy.  Numerical results reported on VOT2016 data set demonstrate that IDPF-RP provides about 7\% higher success rate compared to TCNN \cite{tcnn}, the top tracker of VOT2016. In \cite{tavot} it is shown that false alarm ratio of two-stage deep object detectors, in particular Mask R-CNN, can be reduced by including the region proposal alignment scheme introduced in \cite{GURKAN} into the region proposal network architecture.

Although the trackers summarized above achieve considerably high tracking accuracy, their performance may significantly decrease under scale changes and occlusion.  This is because  the tracked object is located by a rectangular bounding box (BB) rather than a deformed BB that enables us to track the object or a part of it more accurately. To alleviate this drawback,  recent  trackers propose either a part-based tracking by rectangular BBs or tracking by deformed BBs. \cite{gated} introduces  a  deformable  convolutional  layer  that  generates  new features to enrich the part-based representation of the target appearance.
Fusion of the deformable and conventional CNN features is achieved by a gated fusion scheme that monitors how the captured variations affect the original appearance.   
\cite{gated} also updates the tracker with online re-training by using positive and negative samples collected  around the estimated target location.
Reliable Patch Tracker \cite{pat} runs an individual PF to track each object part. Distribution of the reliable patches among the target object proposals generated by each of the particle filters is formulated as a function of confidence level and similarity score of the patch. The confidence level of a sampled patch is specified as the peak-to-side-lobe-ratio of the patch achieved by the correlation filter \cite{corr}. The similarity score is defined as the distance between the trace of the tracked object and the particle. Final BB of the tracked object that maximizes the reliable patch distribution is estimated as the weighted average of the reliable BBs proposed by all particles. Despite they highly improve robustness to occlusion and scale changes, high computational complexity is the main drawback of the part-based object tracking methods. This is why the trackers summarized in the following paragraph so as the proposed tracker employ the deformed BBs to locate the tracked object.

Motivated by \cite{sparse_r},  \cite{l1} presents a pioneering work on sparse regularization for object tracking based on particle filtering. In particular, \cite{l1} formulates a particle state vector in which the affine motion parameters constitute attributes, hence enables us to track object boundaries with deformed BBs. The object tracking is formulated as estimating the state of the object which is represented  by a dictionary with a minimum reconstruction error. L1APG \cite{L1APG} extends this model by adding an occlusion detection scheme in which energy of the trivial components of the dictionary contributes to the tracking. Although a slow dictionary update scheme is proposed in \cite{L1APG,l1}, their robustness to object appearance changes is low. \cite{rw} applies a similar one-by-one dictionary update scheme, but the tracked object BB is estimated by a ML detector based on the weights assigned to local image patches that constitute the target object patch. The presented local sparse appearance model improves tracking of deformed object BBs but is not robust to abrupt appearance changes. To overcome this problem in \cite{erdem} a  dictionary update scheme guided by Faster R-CNN deep object detector and completely updates the dictionary is proposed. In particular, the proposed tracker, L1Dpct, monitors the reconstruction error, energy represented by the trivial dictionary code words, and sparsity of the samplers, and the dictionary update is activated  when all metrics reach up to the predefined thresholds for a number of successive video frames. Although the formulated dictionary update scheme significantly improves robustness to abrupt appearance changes, L1Dpct needs fine tuning of the thresholds which is not easy. Nevertheless development of a tracker that is robust to object appearance changes and object scale changes arised from high motion is still a challenging problem.

In this paper, we propose an efficient tracking algorithm  that is capable of  timely updating the target object model and enables us to effectively track object of interest by deformed BBs.  The proposed tracker, motion guided $l$1 norm deep particle filter (L1DPF-M), is designed based on a sparse particle filter tracker guided by a deep object detector. To accurately locate the object of interest we use deformed BBs extracted by employing the instance segmentation masks provided by Mask R-CNN deep object detector. Differ from the previously proposed methods \cite{erdem,Filiz} that fuse the final decision of the tracker and detector, in this work, we introduce an observation model that enables us to enforce consensus between the tracker and detector thus significantly improves the localization capability of the tracker.  Furthermore robustness to scale changes is significantly improved with the formulated new state vector that allows tracking by the deformed object BBs.
Moreover, the proposed dictionary update scheme improves robustness to abrupt appearance changes.

\section{Theoretical Background}
\label{theoretical}

 In the following subsections we briefly describe $l$1
 Accelerated Proximal Gradient (L1APG) \cite{L1APG} method which is accepted as the baseline $l$1 tracker and Mask Region-based Convolutional Neural Network (Mask R-CNN) \cite{MaskRCNN}, the deep object detector used in this work for instance segmentation.

\subsection{Tracking by Sparse Object Representation} 
\label{sta}

Sparse representation at time $t$ aims to model the candidate target object RoI, $\mathbf{Y}_t$, by a linear combination of a set of significant code words that constitutes a dictionary with $ n $ vectors, denoted by the matrix $\mathbf{S}_t \in \mathbb{R}^{d\times n}$. The dictionary is initialized by collecting $n$ patches cropped
within a one-pixel neighborhood of $\textbf{Y}_t$ and to fix the dimension of the vector space, each of the code words so as the target RoI are converted to a $d$ dimensional vector ($d>>n$) by down-sampling or up-sampling. Considering each pixel of the target RoI can be effected from occlusion or noise,  $d$ trivial code words, each with a single non-zero elements, are also added into the dictionary where $\mathbf{I} \in \mathbb{R}^{d\times d}$ stands for the matrix of trivial code words \cite{l1}. Eq.\ref{eq:sparse} formulates $\textbf{y}_t$, the sparse representation of the candidate target object RoI in vector form at video frame $t$; 

\begin{equation}
\label{eq:sparse}
\mathbf{y}_t\simeq\mathbf{S}_{t}.\mathbf{a}_\textrm{S}+\mathbf{I}.\mathbf{a}_\textrm{I}
\end{equation}
where $\textbf{a}_\textrm{I} $  and  $\textbf{a}_\textrm{S}=[a_{\textrm{S},1},a_{\textrm{S},2}...a_{\textrm{S},n}] $ , $ \forall \textbf{a}_\textrm{S} \geq \mathbf{0} $  respectively denote the trivial and significant coefficient vectors. 

Since Eq.\ref{eq:sparse} does not have a unique solution for  the coefficient vector $\text{a}_t=[\text{a}_\textrm{S},\text{a}_\textrm{I}]$, a sparse solution  can be obtained by solving the problem with Accelerated Proximal Gradient Approach \cite{L1APG} as a $l$1-regularized least square estimation problem such that;

\begin{equation}
\label{eq:l1reg}
\mathbf{c}_{t} = \textrm{argmin}_{\mathbf{a}_{t}} ~\ \|\mathbf{y}_{t}-\mathbf{D}_{t}\mathbf{a}_t\|_2^2+ \lambda {\|\mathbf{a}_t\|}_{1} + \mu_{t} {\| \mathbf{a}_\textrm{I} \|}_{2}^{2}  
\end{equation}
where ${\textbf{c}}_{t}=[\textbf{c}_\textrm{S},\textbf{c}_\textrm{I}]$ is the estimated coefficient vector corresponding to $\textbf{y}_t$, $\textbf{D}_{t}=[\textbf{S}_{t},\textbf{I}]$ is the complete dictionary, ${\| \cdot \|}_{1}$  and ${\|  \cdot \|}_{2}$ respectively  denote $l$1 and $l$2 norms, and  $\lambda$ is the regularization coefficient.  Here $\mu_{t}$ is the smoothing parameter controlling contribution of the energy of the trivial code words.

Under sequential Monte Carlo framework, each particle $i$ points to a candidate object RoI,  $\textbf{y}_t^i$ $t$, $i=1,..,N$, for the video frame , where $N$ denotes the number of particles. Hence a coefficient vector $\textbf{c}_t^{i}$ is estimated for each particle by using Eq.\ref{eq:l1reg}. Moreover, $\textbf{v}_t^i=\{\alpha_1,\alpha_2,\alpha_3,\alpha_4,o_1,o_2\}$,  the state vector that points out $\textbf{y}_t^i$, is modeled  by a set of affine transformation  parameters where the first four are deformation parameters and the last two are translation parameters \cite{L1APG}. Particle sampling has been performed based on the state transition model shown in Eq.\ref{state} where each $\mathcal{N}_k$ is drawn from a zero mean independent Gaussian distribution $\mathcal{N}(0,\sigma_k^2)$.

\begin{equation}
\mathbf{A}_t^i=
      \begin{vmatrix}
    \alpha_1&\alpha_2&o_1\\
\alpha_3&\alpha_4&o_1\\

\end{vmatrix}
_{t}^i=
    \begin{vmatrix}
    \alpha_1&\alpha_2& o_1\\
\alpha_3&\alpha_4& o_1\\

\end{vmatrix}
_{t-1}^i+
    \begin{vmatrix}
\mathcal{N}_1&\mathcal{N}_2&\mathcal{N}_{5}\\
\mathcal{N}_3&\mathcal{N}_4&\mathcal{N}_{6}\\
\end{vmatrix}
\label{state}
 \end{equation}

  The candidate object BB corresponding to the sampled ROI, $\textbf{y}_t^i$, is represented in the matrix form as $ \mathbf{B}_t^i=\{(e_1,b_1)^T,(e_2,b_2)^T,(e_3,b_3)^T,(e_4,b_4)^T\} \in \mathbb{R}^{2\times 4}$, where each (\textit{e},\textit{b})  denotes a corner pixel of the BB and 
it is formulated by  Eq.\ref{eq:oldtrans},

\begin{equation}
\mathbf{B}_t^i = \mathbf{A}_t^i \hspace{0.08 cm}  \textrm{x}  \hspace{0.08 cm} \mathbf{Ref}
\label{eq:oldtrans}
\end{equation} 
where $\mathbf{Ref} $ denotes the corner pixel coordinates of an origin centered reference RoI having size of $\sqrt{d} \textrm{x}\sqrt{d}$. BBs surrounding the candidate object patches are propagated through the frames by affine transformation, hence using this reference ensures consistency between the transformations of each particle as every transformation is applied to the same reference.

The observation likelihood of the video frame, $\mathbf{z}_t$, is formulated to reflect the similarity  between a candidate object RoI  $\textbf{y}_t^i$ and its sparse representation. Hence, it is defined in terms of the minimum reconstruction error achieved by the significant code words as in Eq.\ref{eq:observationmodels}, where $\alpha$ controls the shape of Gaussian kernel,  $L$ is a normalization factor that guarantees $\sum_{i=1}^{N} p_{\scriptscriptstyle PF}(\mathbf{z}_{t}|\textbf{v}_{t}^i) = 1$ and $\textbf{c}_\textrm{S}^i$ is significant part of the coefficient vector estimated by Eq.\ref{eq:l1reg}.

\begin{equation}
\label{eq:observationmodels}
p_{\scriptscriptstyle PF}(\mathbf{z}_{t}|\textbf{v}_{t}^i) = \dfrac{1}{L}e^{-\alpha \ {\| \textbf{y}^i_{t}-\textbf{S}_t.\textbf{c}_\textrm{S}^i \|}^2_{2}},  \ \hspace{0.2 cm} i=1,\cdots,N 
\end{equation}

The candidate object RoI which maximizes the observation likelihood, in other words which minimizes the reconstruction error, is chosen as the estimated object RoI at time $t$. 
Since the observation likelihood specifies the weight of the corresponding particle, the candidate object RoI maximizing Eq. \ref{eq:observationmodels} is decided as the one pointed by the particle with the highest weight. 
Hence the candidate RoI  ${\textbf{y}}_{t}^*$ corresponding to the state vector ${\textbf{v}}_{t}^*$ estimated by Eq. \ref{eq:highestobservationlikelihoodofx} is  declared as the tracked object.

\begin{equation}
\label{eq:highestobservationlikelihoodofx}
{\textbf{v}}_{t}^*= \textrm{argmax}_{\textbf{v}_{t}^i}  p_{\scriptscriptstyle DPF}(\mathbf{z}_{t}|\textbf{v}_{t}^i)
\end{equation}

In order to improve the representation capability of the dictionary a slow update scheme is also  proposed in \cite{L1APG}. In particular,  whenever the similarity between the tracked object RoI and the existing code words remains lower than a pre-specified similarity threshold, the code word having the smallest coefficient is replaced by the tracked RoI that yields a dictionary update.  Since these slow updates do not ensure robustness to appearance changes, several alternative update schemes are proposed in the literature \cite{rw}. The new dictionary update scheme proposed in this paper is formulated in Section \ref{newobsv}.

\subsection{Object Detection by Mask R-CNN}
\label{maskdet}
Most of the existing object detectors including deep detectors \cite{faster,yolo,ssd} predict a rectangle  BB to localize the target object. Unlike these methods, Mask R-CNN  not only provides a well localized rectangle bounding box but also an instance segmentation mask for each detected object  \cite{MaskRCNN}. In our tracking model, these segmentation masks are employed to track the  object RoIs with deformed BBs that enables us to localize the tracked object more accurately, especially for the transformed objects. This is why we integrate Mask R-CNN as the discriminative object detector into the our tracking model.
\begin{figure*}[t]
\centering
\includegraphics[width=4 in]{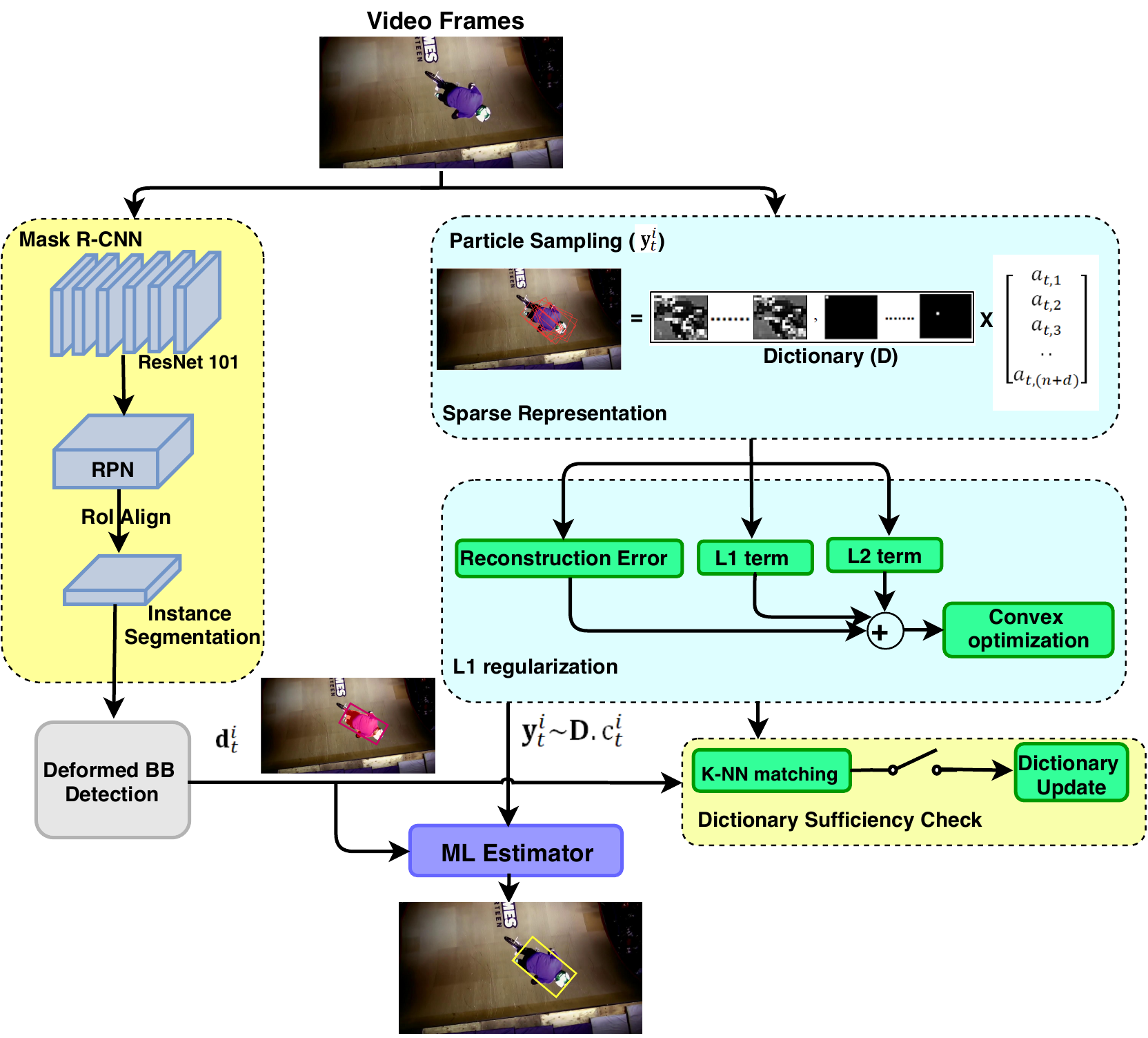}
\caption{Main processing blocks of the proposed L1DPF-M tracker.}
\label{fig:pro}
\end{figure*}

Mask R-CNN architecture consists of three major layers; a ResNet101 \cite{resnet} backbone for feature extraction, a region proposal network (RPN) \cite{faster} that generates candidate object BBs with their corresponding objectness scores and a head architecture that performs multi-class classification, bounding box regression and instance segmentation. 
RPN takes the feature maps produced by ResNet101 as an input and outputs a set of rectangular object proposals, with their binary objectness scores. These RoIs are resized to a fixed dimension by RoIAlign. Differ from RoIPooling \cite{fastrcnn} that uses quantization,  RoIAling applies bilinear interpolation \cite{stn}  on the input features that increases the accuracy. The head architecture of Mask R-CNN has two main branches, one is bounding box regression and classification and the other one is instance segmentation as known as the mask branch. 

Eq.\ref{eq:Head_loss} denotes the loss function minimized at the training stage of  Mask R-CNN deep detector;
\begin{equation}
\small{L_{Head} = \frac{1}{R} \sum_{i=1}^R
L_{cls}(\textbf{k}^i, \textbf{k}^{i*}) + L_{reg}(\boldsymbol{v}^{i}, \boldsymbol{v}^{i*}) + L_{mask}(\mathbf{M}^{i}, \mathbf{M}^{i*})}
\label{eq:Head_loss}
\end{equation}
where $R$ denotes the total number of RoIs produced by RPN to be used during the training, $L_{cls}$  is the categorical cross-entropy function between the predicted ($\textbf{y}^i$) and ground truth ($\textbf{y}^{i*}$) class labels for the $i^{th}$ RoI.  $L_{reg}$ is the smooth $l$1 loss function for the predicted state vector $\boldsymbol{v}^i$ and ground truth $\boldsymbol{v}^{i*}$. $L_{mask}$ is the binary cross-entropy function between the predicted segmentation mask $\mathbf{M}^{i}$ and ground truth $\mathbf{M}^{i*}$.

In order to be consistent with the state-of-the-art training efforts, we have used the model file provided at \url{https://github.com/matterport/Mask\_RCNN/releases/tag/v1.0} for inference. In particular the instance segmentation masks generated by the model file guide our tracker as it is formulated in subsection \ref{newobsv}.

\section{L1DPF-M: The Proposed Object Tracker}
\label{proposed}

We propose a tracking-by-detection method named as motion guided $l$1 norm deep particle filter (L1DPF-M), which is designed to improve robustness to appearance changes arised from occlusion, high motion and scale changes. This is achieved by integrating $l$1 tracker with a deep object detector.
Main processing blocks of L1DPF-M described in this section are illustrated at Figure \ref{fig:pro}. A target object RoI is specified at the first frame of tracking by a BB referred as the target object BB and the initial dictionary code words are generated from the target object RoI. The deep object detector, Mask R-CNN, and $l$1 regularized sparse particle filtering are simultaneously activated at the successive frames. 
 The particle sampling mechanism included in the sparse representation block generates the candidate object BBs and each of them is reconstructed by the initial dictionary code words. Optimal weights of each candidate region are iteratively estimated by $l$1 regularization that minimizes the reconstruction error under the non-negative weight constraint (Eq.\ref{eq:l1reg}). The candidate object RoI having the minimum reconstruction error is fed into a ML estimator to be processed together with the candidate object RoI detected by Mask R-CNN instance segmentation branch. The ML estimator locates the final tracked object BB that maximizes the observation model of L1DPF-M introduced in subsection \ref{newobsv}. Moreover, the novel dictionary update scheme of L1DPF-M that monitors sufficiency of the existing dictionary and activates a complete dictionary update mechanism, if necessary, is formulated in subsection \ref{targetupdate}.

\subsection{Observation Model of L1DPF-M}
\label{newobsv}
Observation model of the conventional $l$1 tracker given by Eq.\ref{eq:observationmodels} reflects the similarity between a candidate object RoI pointed by a particle and its reconstruction by the significant code words. In particular, the reconstruction error shown as an exponential term in Eq.\ref{eq:observationmodels} provides  a measure to check the sufficiency of the dictionary at frame $t$. The main drawback of the formulated generative tracker is it fails in high motion video sequences where abrupt scale and pose changes encountered at the objects. To avoid drifts from the target model, we formulate a novel observation model which forces consensus between the detector and the tracker. The new observation model for the proposed tracker is formulated in Eq.\ref{eq:observationmodels2},

\begin{equation*}
\label{aa}
p_{\scriptscriptstyle DPF}(\mathbf{z}_{t}|\mathbf{v}_{t}^i) = p_{\scriptscriptstyle PF}(\mathbf{z}_{t}|\mathbf{v}_{t}^i) p_{\scriptscriptstyle DL}(\mathbf{d}_{t}|\mathbf{y}_{t}^i)\\ 
\end{equation*}
\vspace{-0.6 cm}
\begin{equation}
\label{eq:observationmodels2}
\hspace{2.15in}=\dfrac{1}{K}e^{-\alpha (\ {\| \mathbf{y}^i_{t}-\textbf{S}_t.\mathbf{c}_\mathbf{S}^i \|}^2_{2}+\ {\|\mathbf{d}_t-\mathbf{y}_t^i  \|}^2_{2})},  \ i=1..N \hspace{0.3in} 
\end{equation}\\

\hspace{-0.2in}where $\mathbf{d}_t$ is the object patch detected by Mask R-CNN instance segmentation layer and than shaped by the deformed BB detection block at time $t$, $K$ is a normalization parameter.

\begin{figure}[ht]
 \centering
  \begin{subfigure}[ht]{2 in}
   \includegraphics[width=\linewidth]{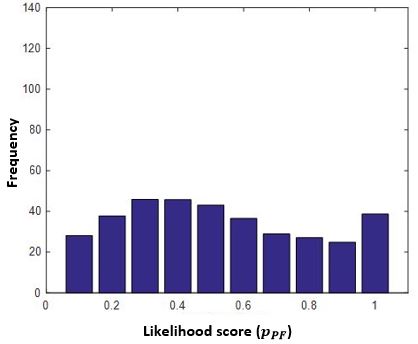}
    \caption{}
  \end{subfigure}
  \begin{subfigure}[ht]{2   in}
    \includegraphics[width=\linewidth]{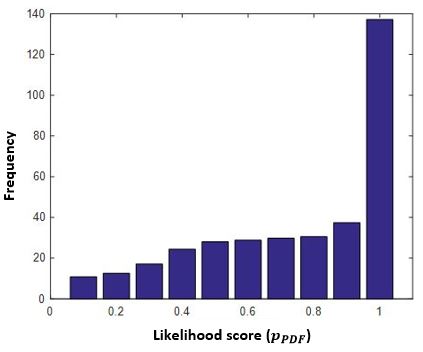}
   \caption{} 
  \end{subfigure}
 \caption{ Histogram of the likelihood scores obtained on VOT2016 data set (a) by using the baseline observation model  and (b) by using the proposed observation model.}
 \label{fig:im3}
\end{figure}

The likelihood score formulated by Eq.\ref{eq:observationmodels2} corresponds to the weight of the $i$th particle thus indicates contribution of the particle on the final estimation where $p_{\scriptscriptstyle PF}(\mathbf{z}_{t}|\textbf{v}_{t}^i)$ denotes the likelihood term coming from the conventional observation model of particle filter and $p_{\scriptscriptstyle DL}(\mathbf{d}_{t}|\textbf{y}_{t}^i)$ models the contribution of the deep object detector.
In particular, $p_{\scriptscriptstyle DL}(\mathbf{d}_{t}|\textbf{y}_{t}^i)$ hence the additional error term shown at the exponent of Eq.\ref{eq:observationmodels2} adaptively controls similarity between the detected object patch and the patch pointed by $i^{th}$ particle. Higher similarity increases the likelihood score so as the probability of the BB pointed by $i^{th}$ particle being the final estimation.
Histograms of likelihood scores obtained on VOT2016 data set are respectively shown at Figure \ref{fig:im3}(a) and Figure \ref{fig:im3}(b). It can be concluded that the proposed observation model enforces agreement between the PF tracker and the deep object detector that yields more discriminative representations (Figure \ref{fig:im3}(b)) whereas the histogram corresponding to the conventional method (Figure \ref{fig:im3}(a)) is almost uniform.

The candidate RoI  ${\textbf{y}}_{t}^*$ corresponding to the state vector ${\textbf{v}}_{t}^*$ estimated by Eq.\ref{eq:obsvlklhd} is  declared as the tracked object.

\begin{equation}
\label{eq:obsvlklhd}
{\textbf{v}}_{t}^*= \textrm{argmax}_{\textbf{v}_{t}^i}  p_{\scriptscriptstyle DPF}(\mathbf{z}_{t}|\textbf{v}_{t}^i)
\end{equation}

\subsection{Deformed Object Bounding Boxes}
\label{defBB}

\begin{figure}[ht]
 \centering
  \begin{subfigure}[ht]{2.6 in}
   \includegraphics[width=\linewidth]{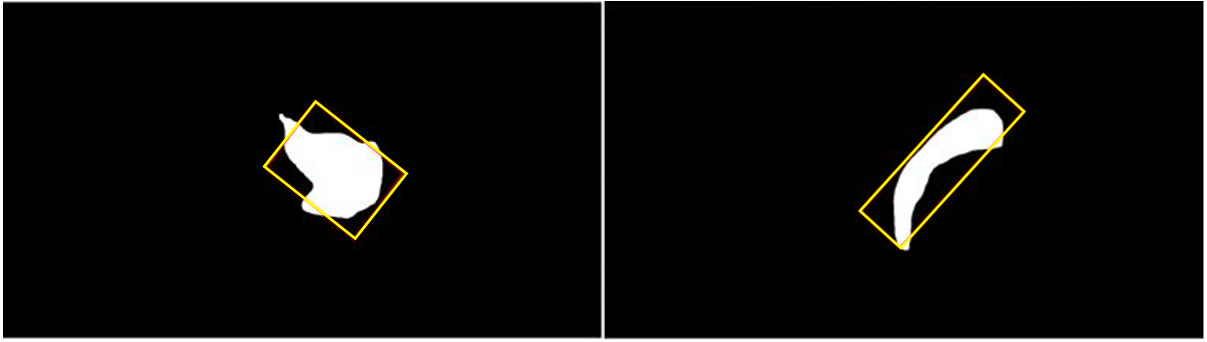}
    \caption{}
  \end{subfigure}
  \begin{subfigure}[h]{2.6 in}
    \includegraphics[width=\linewidth]{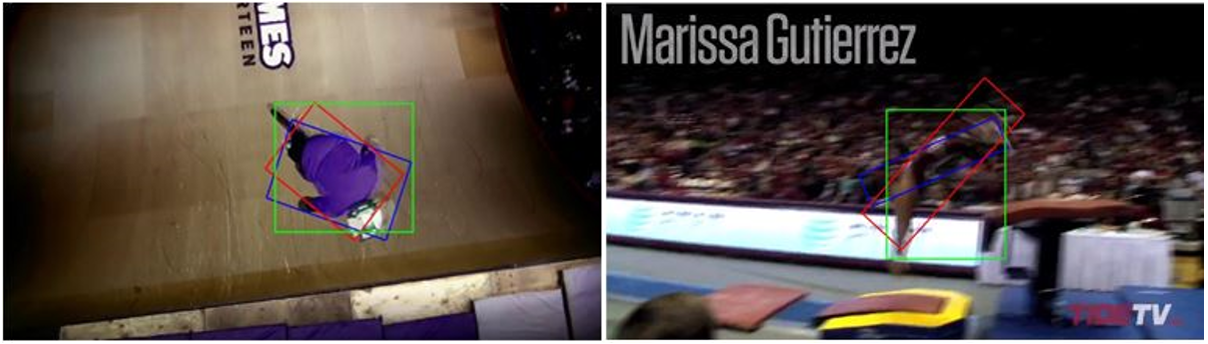}
   \caption{} 
  \end{subfigure}
\caption{ (a) Deformed object BBs (yellow) extracted by using instance segmentation of Mask R-CNN. (b) Deformed BB (red), rectangular BB(green), and ground truth BB (blue) overlaid on a video frame from sequence Bmx (frame no: 58) (left) and  from sequence Gymnastics3 (frame no: 40) (right).}
\label{fig:mask}
\end{figure}

Conventional trackers and detectors estimate the object BB as a 4-D vector $\textbf{b}_t=\{e,b,w,h\}$ where $\{e,b\}$ refers the left top corner of the tracked BB and $\{w,h\}$ respectively denote the width and height of the tracked BB. Consequently the tracked object BB is represented by a rectangular BB (rectBB) that prevents to minimize number of the background pixels included in the tracked object RoI \cite{GURKAN,fastrcnn,tcnn,faster,siam,MLDF}. Furthermore the rectBBs are not feasible for accurate tracking of the objects affected by deformations through the video sequence. Figure \ref{fig:mask} illustrates advantage of using deformed BBs rather than rectBBs. In particular, intersection-of-union (IoU) of the deformed BB and ground truth is calculated as 0.7 and 0.35 for two objects shown at Figure \ref{fig:mask} (b), while it is respectively  decreased to 0.57 and 0.23 for the rectangular BB. Therefore, the ground truth data provided at state-of-the-art benchmarking evaluations are prepared in the form of deformed BBs \cite{VOT2018}. To take the advantage of tracking by the deformed BBs, we introduce a new state vector in Section \ref{proposed2} that enables us to represent the object boundaries with deformed BBs. 
Hence the tracked object BB at frame $t$ is represented by a matrix, $\mathbf{B}_t^i=\{(e_1,b_1)^T,(e_2,b_2)^T,(e_3,b_3)^T,(e_4,b_4)^T\} \in \mathbb{R}^{2\times 4}$, where each $(e,b)$  shows the coordinate of a corner pixel location of the deformed object BB. Correspondingly  the detected object RoI at frame $t$ is denoted by $\mathbf{D}_t$. 
A deformed BB is extracted based on the instance segmentation mask provided by the Mask R-CNN object detector. To achieve this we fit an ellipse onto the instance segmentation mask where the segmentation identifies the label of each pixel for every known object within a frame. Since the labels are instance-aware, the mask corresponding to one specific object is considered as a connected component in a 2D image plane. Thus the smallest BB surrounding the ellipse is taken as the deformed object BB detected by the deep detector. The object patch $\mathbf{d}_t$ shown in Eq.\ref{eq:observationmodels2} refers to the patch surrounded by this deformed BB.

\subsection{Target Update}
\label{targetupdate}

 The proposed observation model formulated by Eq.\ref{eq:observationmodels2} enables us to select the qualified particles more accurately, however it is difficult to design a tracker that is robust to the object appearance changes without updating the target model that corresponds updating the dictionary in our sparse representation scheme. A smooth updating model of L1APG allows replacement of at most one code word per frame depending on a correlation based similarity measure that is not sufficient to prevent drifts from the target model under occlusion and pose changes. In order to alleviate this drawback, we propose a new template set update scheme for L1DPF-M that performs a full target template set update when it is decided that the existing dictionary is not sufficient. 
 
 Target update is activated according to a k-NN matching score defined on the likelihood probabilities. Specifically, if the maksimum likelihood score $p_{\scriptscriptstyle PF}(\mathbf{z}_{t}|\textbf{v}_{t}^i)$ remains within k-NN of the likelihood score set of $p_{\scriptscriptstyle DPF}(\mathbf{z}_{t}|\textbf{v}_{t}^i), i_1,..,N$, we assume that the detector and tracker install a consensus on the detected object BB and  it means the dictionary can still effectively represent the object. Otherwise it is assumed that the tracker and detector are disagree than a dictionary update is initialized with the guidance of the deep detector. L1DPF-M performs the dictionary update by replacing the $n$ significant code words of the existing dictionary with the set constructed by collecting $n$ patches within a neighborhood of the ROI detected by Mask R-CNN.

\begin{figure}[ht]
 \centering
  \begin{subfigure}[ht]{4 in}
   \includegraphics[width=\linewidth]{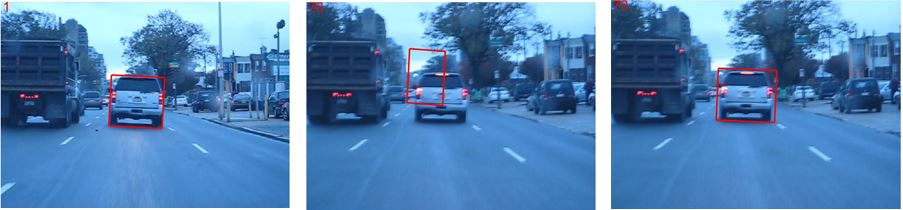}
    \caption{}
  \end{subfigure}
  \begin{subfigure}[ht]{4 in}
    \includegraphics[width=\linewidth]{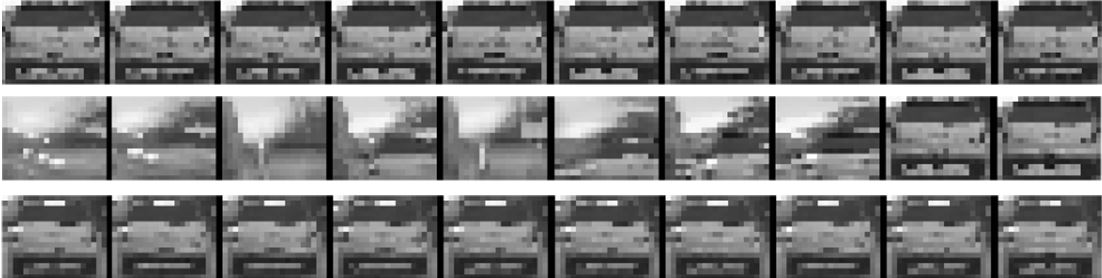}
   \caption{} 
  \end{subfigure}
 \caption{Video sequence car1. (a) left to right- BB visualization, frame no:1 (initialized target BB), frame no:70 (BB tracked by L1APG), frame no:70 (BB tracked by L1DPF-M), (b) top to down- Dictionary visualization,  frame no:1 (initial dictionary), frame no:70 (dictionary of L1APG), frame no:70 (dictionary of L1DPF-M).}
 \label{fig:TU}
\end{figure}
Figure \ref{fig:TU} illustrates the dictionary constructed by L1DPF-M and by the slow update scheme of L1APG. Both methods take the same target object BB given at the first frame thus the initial dictionary code words are the same.  It is clearly observable from Figure \ref{fig:TU}(b) that the dictionary of L1DPF-M is effectively model the target object thus accurately localize the tracked object at frame number 70, however the dictionary of L1APG drifts from the target object thus provides a poorly localized object BB (Figure \ref{fig:TU}(a)).

\section{Enhanced Robustness via Motion-based State Transitions }
\label{proposed2}

The state vector formulated in the previous section that assumes the 6 affine motion parameters independently sampled by Eq.\ref{state} enables us to track the deformed object BBs thus improves robustness to appearance changes.
 However, since the deformations arise from scale change, rotation and shearing
are highly correlated, it is much more effective to use a state vector that reflects this dependency. Therefore in this work we propose a new state vector  $\textbf{v}_t^i$ of $i^{th}$ particle at time $t$,
 \begin{equation}
\textbf{v}_t^i=(\theta,o_1,o_2,s_1, s_2, sh_1,sh_2)_{t}^i
 \end{equation}
where $\theta$ refers the rotation parameter, $ o_1,o_2$ denote translation, $ s_1, s_2 $ refer scaling and $ sh_1,sh_2 $ denote the shearing parameters, respectively in $y$ and $x$ direction. 
Each of the parameters are drawn from independent Gaussian distributions as it is formulated in Eq.\ref{newsv},

 \begin{equation}
    \textbf{v}_t^i= \textbf{v}_{t-1}^i+\mathcal{N}(0,\mathbf{\sigma}_k^2)
     \label{newsv} \hspace{0.4 cm} k=1...7.
 \end{equation}

The new state vector allows to form separate translation, rotation, scaling and shearing components. Because the same parameters can be reached through different orders of transformations, the affine transformation between consecutive frames are estimated by applying the respective translations to the previous states in order. This enables the system to track each motion parameter individually that enables us to much more accurately interpret the type of motion from the estimated parameters. 
Eq. \ref{state2} shows the affine transformation matrix $\mathbf{A}_t^i$ for $i^{th}$ particle at time $t$,
\begin{equation}
\mathbf{A}_t^i =\begin{vmatrix}
    \alpha_1&\alpha_2&o_1\\
\alpha_3&\alpha_4&o_1\\

\end{vmatrix}_t^i
= \mathbf{T}_t^i*\mathbf{R}_t^i*\mathbf{Sh}_t^i*\mathbf{Sc}_t^i\\
\label{state2}
 \end{equation}
where Eq.\ref{ma1} and \ref{ma2} formulate the rotation (R), translation (T), scaling (Sc) and shearing (Sh) matrices.

\begin{equation}
\mathbf{R} = 
 \begin{vmatrix}
cos\theta&sin\theta&0\\
-sin\theta&cos\theta&0\\
0&0&1\\
\end{vmatrix}_t^i
\hspace{0.5cm} 
\mathbf{T} = 
 \begin{vmatrix}
1&0&o_1\\
0&1&o_2\\
0&0&1\\
\end{vmatrix}_t^i
\label{ma1}
\end{equation} 

\begin{equation}
\mathbf{Sc} = 
 \begin{vmatrix}
s_1&0&0\\
0&s_2&0\\
0&0&1\\
\end{vmatrix}_t^i
\hspace{0.5cm}
\mathbf{Sh} = 
 \begin{vmatrix}
1&sh_1&0\\
sh_2&1&0\\
0&0&1\\
\end{vmatrix}_t^i
\label{ma2}
\end{equation} 
Here, each matrix is formed to represent the related transformation for that component. For example, $\mathbf{R}$ matrix represents a rotation of ROI by $\theta $ degrees with respect to the origin and $\mathbf{T}$ matrix models a translation of ROI in each dimension by $ o_1,o_2 $ pixels and so on. The affine transformation matrix can be rewritten as in Eq.\ref{state3}. Differ from Eq.\ref{state}, Eq.\ref{state3} clearly illustrates dependency between the transformation parameters.

\begin{equation}
\mathbf{A}_t^i =  \begin{vmatrix}
   s_1*(cos\theta+sh_2*sin\theta)&s_2*(sin\theta+sh_1*cos\theta)&o_1\\ 
 s_1*(sh_2*cos\theta-sin\theta)&s_2*(cos\theta+sh_1*sin\theta)&o_2\\
\end{vmatrix}_t^i
\label{state3}
 \end{equation}

Similar to the conventional method, a candidate RoI pointed  by  each  particle  is  found  by multiplying a reference region coordinate matrix $(\mathbf{Ref})$ with the affine transformation matrix.
Moreover, in order to keep the center of the
tracked region at the origin, all the transformations are modeled with respect to the
center of the object. Furthermore to preserve the alignment of rigid object points the order of transformations is fixed as scaling, shearing and then rotation.

\begin{figure}[ht]
 \centering
  \begin{subfigure}[ht]{3.5 in}
   \includegraphics[width=\linewidth]{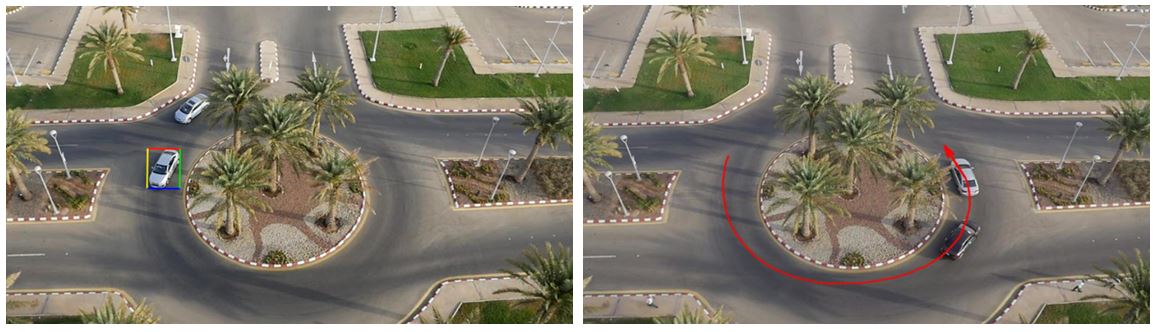}
   \subcaption{}
  \end{subfigure}
   \centering
  \begin{subfigure}[ht]{3.5 in}
   \includegraphics[width=\linewidth]{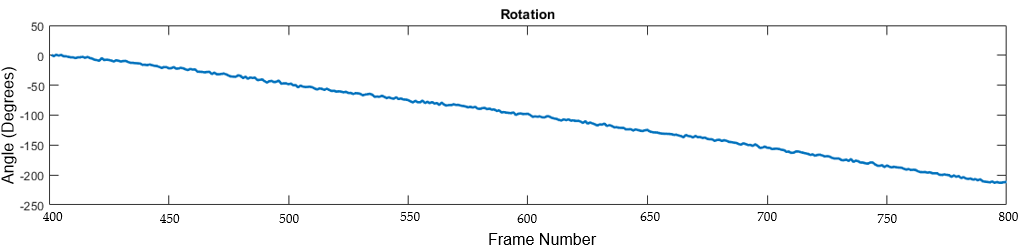}
      \subcaption{}
  \end{subfigure}
 \caption{(a) Video sequence car7, frames 400 and 800. (b) Change of the estimated rotation parameter through 400 video frames.}
 \label{fig:car}
\end{figure}

When the camera is not moving or its motion is compensated, the proposed state vector also makes it possible to interpret the motion of the object as rotation, scaling and translation via observing each parameter individually. Figure \ref{fig:car}(a) shows example frames where the object motion is rotation and Figure \ref{fig:car}(b) demonstrates the estimated rotation parameter $\theta$ throughout the video. Here, green BB in the first frame denotes GT and in the second frame the object trajectory is plotted by the red line. It is clear that $\theta$ parameter is initially set to $0$ in the first frame  and the tracker estimates the rotation of object with respect to the reference. 

\section{Performance Evaluation}
\label{perf}
This section first reports the results of an ablation study that demonstrates individual contribution of the novel observation model that fuses the tracker and detector, the new motion based state vector that enables tracking deformed objects, and the dictionary update scheme that improves robustness to appearance changes. Then the tracking results achieved by L1DPF-M are reported compared to the state-of-the-art trackers. 

The proposed tracker, L1DPF-M, is evaluated on commonly used Visual Object Tracker Benchmark (VOT) 2016 and 2018 datasets \cite{VOT2016,VOT2018}. 
We used  37 of the VOT2016 videos and 31 of the VOT2018 videos that include object classes learned by the released model of Mask R- CNN trained on COCO \cite{coco}. Mask R-CNN object detection results are obtained in TensorFlow by modifying the code available on ”\url{{https://github.com/matterport/Mask\_RCNN}}” and L1DPF-M is implemented in MATLAB.
Experiments and evaluations are conducted with Intel Core i7 4790 CPU 3.6 GHz and GeForce GTX TITAN X GPU.
 
 Performance reported by using common evaluation metrics,  success plot, accuracy and robustness. In particular, success plot is distribution of success rate versus Intersection of Union (IoU) threshold where the success rate is the ratio of the successful frames. Accuracy (A) is formulated as the average IoU calculated over all successfully tracked video frames and robustness (R) measures how many times the tracker drifts off the target. All tracking results are accessible at \url{https://github.com/msprITU/L1DPFM}.

\subsection{Ablation study}
\label{ablat}

\begin{figure}[ht!]
 \centering
  \begin{subfigure}[ht]{2.5 in}
   \includegraphics[width=\linewidth]{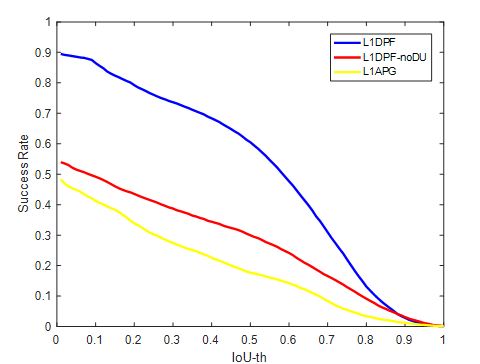}
   \caption{}
   \end{subfigure}
   \begin{subfigure}[ht]{2.5 in}
   \includegraphics[width=\linewidth]{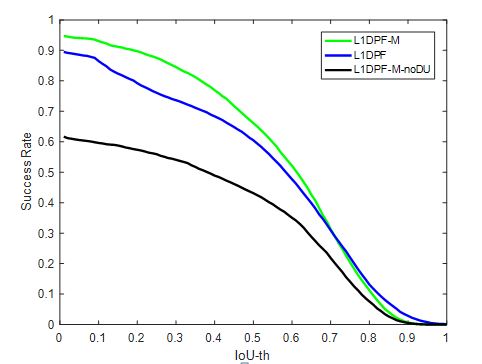}
   \caption{}
   \end{subfigure}
 \caption{(a) Improvement achieved by the proposed observation model  and the dictionary update scheme. (b) Impact of the new state vector on tracking.}
 \label{fig:votovT}
\end{figure}

In order to evaluate improvement achieved by the new observation model, we report the  success rates obtained on VOT2016 data set. Figure \ref{fig:votovT} (a) reports the mean success rates obtained over 37 video sequences at different IoU threshold values where IoU-th=0.5 means overlapping between the tracked object BB and the ground truth BB is equal to or greater than 50\%.  L1APG (yellow line) stands for the work that can be considered as the baseline \cite{L1APG} since it employs the observation model formulated by Eq.\ref{eq:observationmodels} and does not apply the target update proposed in subsection \ref{targetupdate}. L1DPF (blue line) indicates the L1DPF-M with conventional state vector, L1DPF-noDU (red line) refers the L1DPF without dictionary update. It is observable that the fusion of deep detector and particle filtering improves object localization accuracy hence the new observation model increases the success rates at all IoU-th values. However, because of the occlusion and abrupt appearance changes, success rates are not satisfactory. Success rates of L1DPF (blue line) demonstrate  improvement achieved by inclusion of the deep detector guided dictionary update along with the new observation model. It can be concluded that the dictionary update significantly increases the success rates especially at low IoU-th values. In order to demonstrate impact of the new state vector on tracking performance, success rates achieved by L1DPF-M that only replaces the state vector of L1DPF with the new state vector formulated by Eq.\ref{newsv} are reported at different IoU-th values. Figure \ref{fig:votovT} (b) illustrates L1DPF-M provides 6\% higher success compared to L1DPF. This is because the new state vector enables us to track deformed object BBs more accurately. Moreover, without dictionary update (L1DPF-M-noDU) success of L1DPF-M drops below L1DPF that clearly demonstrates impact of the proposed dictionary update scheme that improves robustness to object appearance changes and occlusion.

\begin{figure*}[ht]
 \centering
  \begin{subfigure}[ht]{4.6 in}
 
  \includegraphics[width=\linewidth]{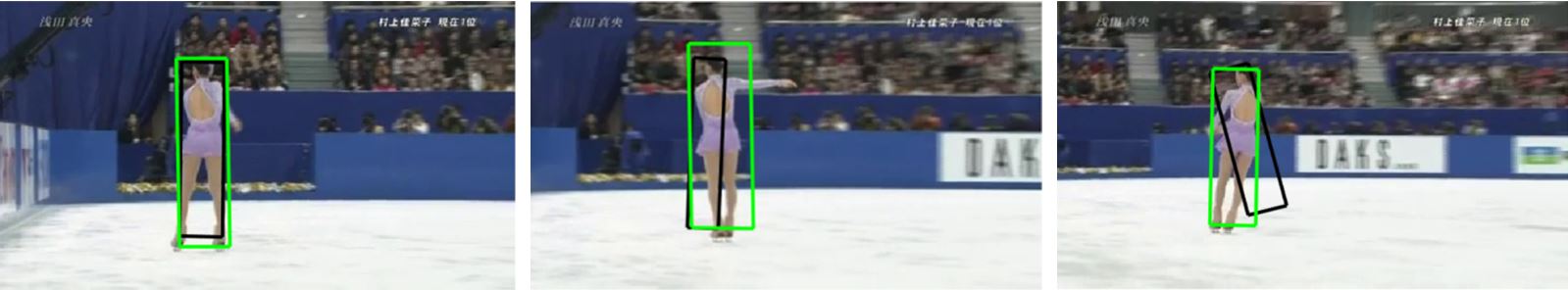}
    \caption{}
  \end{subfigure}

\begin{subfigure}[ht]{4.6in}
  \includegraphics[width=\linewidth]{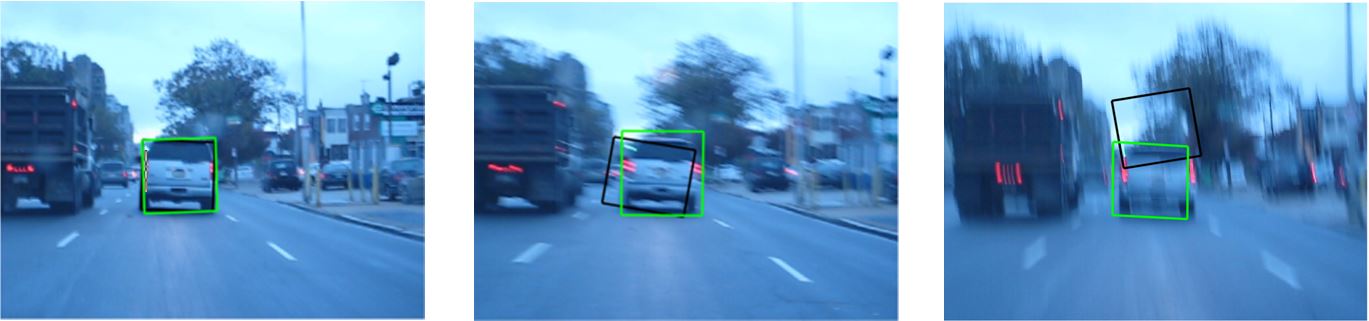}
 \caption{} 
 \end{subfigure}
 \caption{(a) Video sequence: Iceskater1,frame no: 4,15,35. Impact of the introduced observation model. L1DPF-M (green) and L1DPF-M with the  conventional observation model(black). (b) Video sequence: Car1, frame no: 5, 13, 42. Impact of the full dictionary update scheme. L1DPF-M (green), L1DPF-M with the conventional observation model (black). }
 \label{fig:imob}
\end{figure*}

Video frames illustrated in Figure \ref{fig:imob} visually demonstrate impact of the new observation model and the proposed dictionary update scheme on tracking accuracy. As it can be seen from Figure \ref{fig:imob}(a), both L1DPF-M and L1DPF-M with the conventional observation model are initialized with the ground truth at the first frame and provide similar BBs at frame 15. However because of the scale and pose changes, localization accuracy achieved by the conventional model rapidly decreases at the subsequent frames while L1DPF-M remains highly robust as a result of the feedback received from the deep detector. Improvement achieved by L1DPF-M becomes higher at longer sequences because the new observation model encourages re-sampling of particles having more localized BBs that prevents drifts from the target model. 
Figure \ref{fig:imob}(b) illustrates contribution of the proposed target update scheme. Both trackers accurately localize the object at frame 5 however  because of its slow dictionary update scheme, L1DPF-M-noDU drifts the object under motion blur, but L1DPF-M dynamically updates the dictionary based on the proposed full dictionary update scheme thus accurately tracks the object at frames 15 and 35.

\begin{figure*}[ht]
 \centering
  \begin{subfigure}[h]{4.9 in}
   \includegraphics[width=\linewidth]{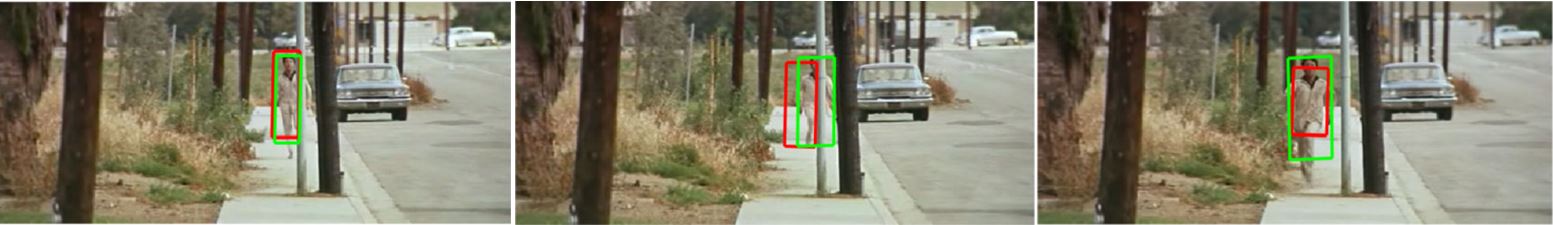}
    \caption{}
  \end{subfigure}
  \begin{subfigure}[ht]{4.9 in}
    \includegraphics[width=\linewidth]{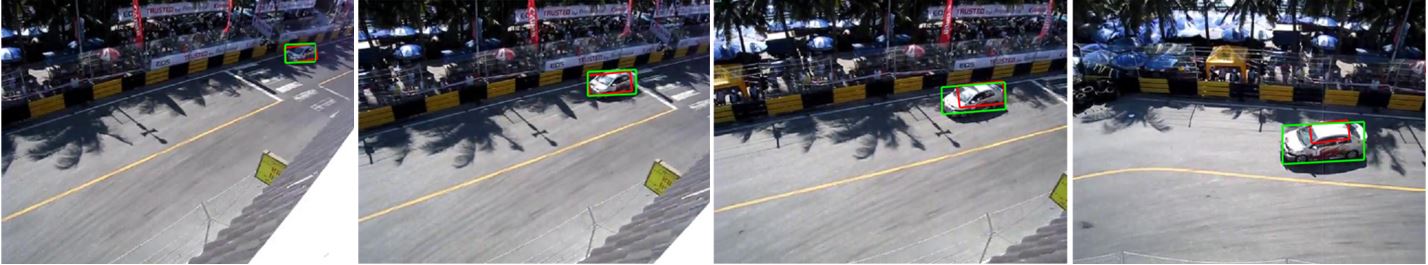}
   \caption{} 
  \end{subfigure}
    \begin{subfigure}[ht]{4.9 in}
    \includegraphics[width=\linewidth]{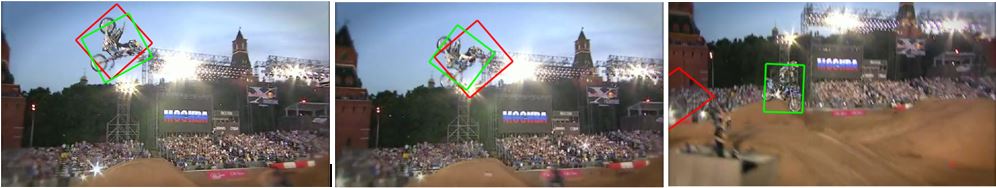}
   \caption{} 
  \end{subfigure}
 \caption{Impact of the state vector. Tracked BB by L1DPF (red BB) and L1DPF-M (green BB) on (a) Video sequence: Graduate, frame no:236,255,323, (b) Video sequence: Racing, frame no:5,30,40,60, (c) Video sequence: Motocross, frame no:17,21,35. }
 \label{fig:8th}
\end{figure*}

In order to demonstrate impact of the state vector on representation of the object motion, we have reported the tracking results obtained by the state vector including uncorrelated affine parameters (L1DPF) and and correlated affine motion parameters (L1DPF-M).
Object BBs tracked by L1DPF and L1DPF-M are illustrated at Figure \ref{fig:8th} where video objects have significant size changes in sequence Graduate and Racing. Furthermore rotation and illumination changes encountered in video sequence Motocross1, abrupt object appearance changes in sequence Racing, and occlusion in sequence Graduate make tracking and detection harder. It can be observable from Figure \ref{fig:8th}(a) and \ref{fig:8th}(b) that L1DPF-M is capable of tracking objects at different scales under occlusion (Figure \ref{fig:8th}(a)) and high illumination as well as appearance changes (Figure \ref{fig:8th}(b)). Whereas L1DPF fails to localize objects at different scales.
Tracking results reported at Figure \ref{fig:8th}(c) make clear advantage of estimating individual motion parameters, specifically, L1DPF-M keeps tracking the rotated motorcycle whereas  L1DPF is not robust to transformations thus drifts from the object of interest.

\begin{figure}[ht]
 \centering
  \begin{subfigure}[h]{2.6in}
   \includegraphics[width=\linewidth]{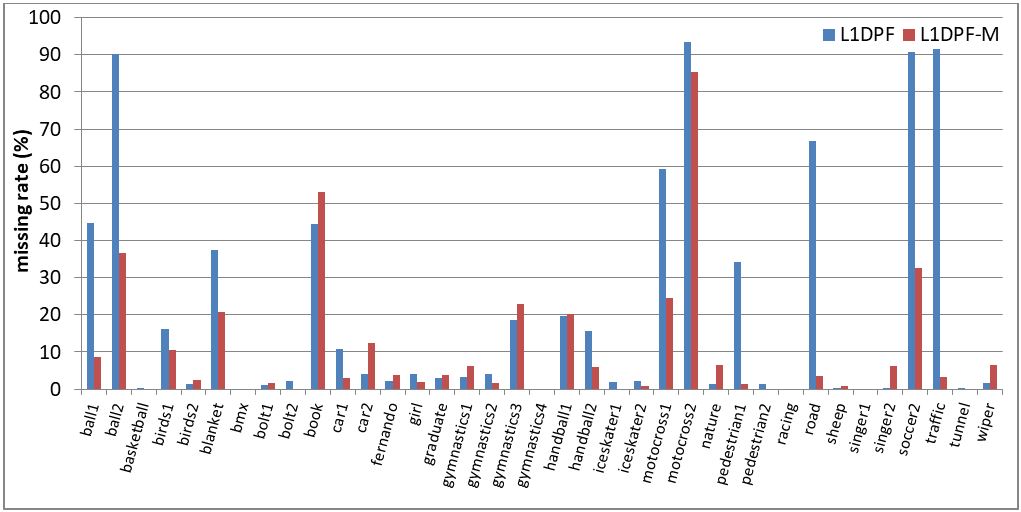}
      \caption{} 
  \end{subfigure}
  \begin{subfigure}[ht]{2.6in}
    \includegraphics[width=\linewidth]{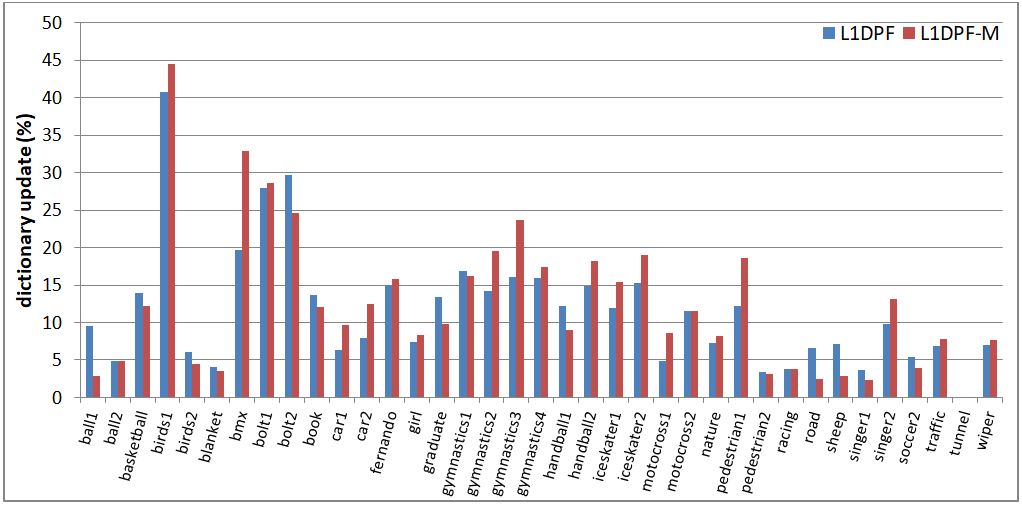}
   \caption{} 
  \end{subfigure}
 \caption{ (a) Sequence based miss detection rates obtained by L1DPF-M and L1DPF on VOT2016  sequences. (b) Percentage of dictionary updates performed by L1DPF and L1DPF-M on VOT2016 sequences.}
 \label{fig:imcom}
\end{figure}

In addition to provide higher success rates, the proposed video object tracker reduces the miss detection rates.
Figure \ref{fig:imcom}(a) reports sequence based miss detection rates obtained on VOT2016 sequences. As it is expected superiority of L1DPF-M becomes more clear depending on the motion content of the video sequence. In particular, for high motion videos L1DPF-M may significantly decrease the miss detection ratio as it is achieved on ball2 sequence where the ratio is reduced from 90\% to 36\%. 
Nevertheless it is clear that L1DPF-M reduces the number of missed frames about 5\% compared to L1DPF. 
Similar to the miss detection rates, the dictionary update rates shown in Figure \ref{fig:imcom}(b) are not uniformly distributed and vary depending on the video content. 
In the average, the number of target updates are very close and respectively reported as 11.7\% and 11.1\% for L1DPF-M and L1PDF. However L1DPF-M provides 6\% higher success rate at IoU-th=0.5.

\subsection{Overall tracking performance} 
\label{overperf}

We have evaluated overall tracking performance of the proposed L1DPF-M tracker compared to the top trackers of VOT2016 benchmarking \cite{VOT2016}, in particular,  TCNN \cite{tcnn}, SSAT and MLDF \cite{MLDF}  as well as the top trackers of VOT2018 benchmarking \cite{VOT2018}, in particular, SiamRPN \cite{siam}, DLSTpp  and MFT. 
Three of these trackers, TCNN,SSAT, and MLDF, employ the deep neural networks, two of them, DLSTpp and MFT, apply discriminative correlation filter and SiamRPN is designed based on the siamese network. TCNN (Tree-CNN)\cite{tcnn} maintains a multiple target appearance model based on CNNs embedded in a tree structure. The DLSTpp (Deep Location-Specific Tracking) which is the top tracker of VOT2018 unsupervised test case, decomposes the tracking problem as classification and localization where the localization is achieved by ECO\cite{eco} and MDNet is used as the classification network \cite{mdnet}.

\begin{figure}[ht]
 \centering
  \begin{subfigure}[ht]{2.4 in}
  \includegraphics[width=\linewidth]{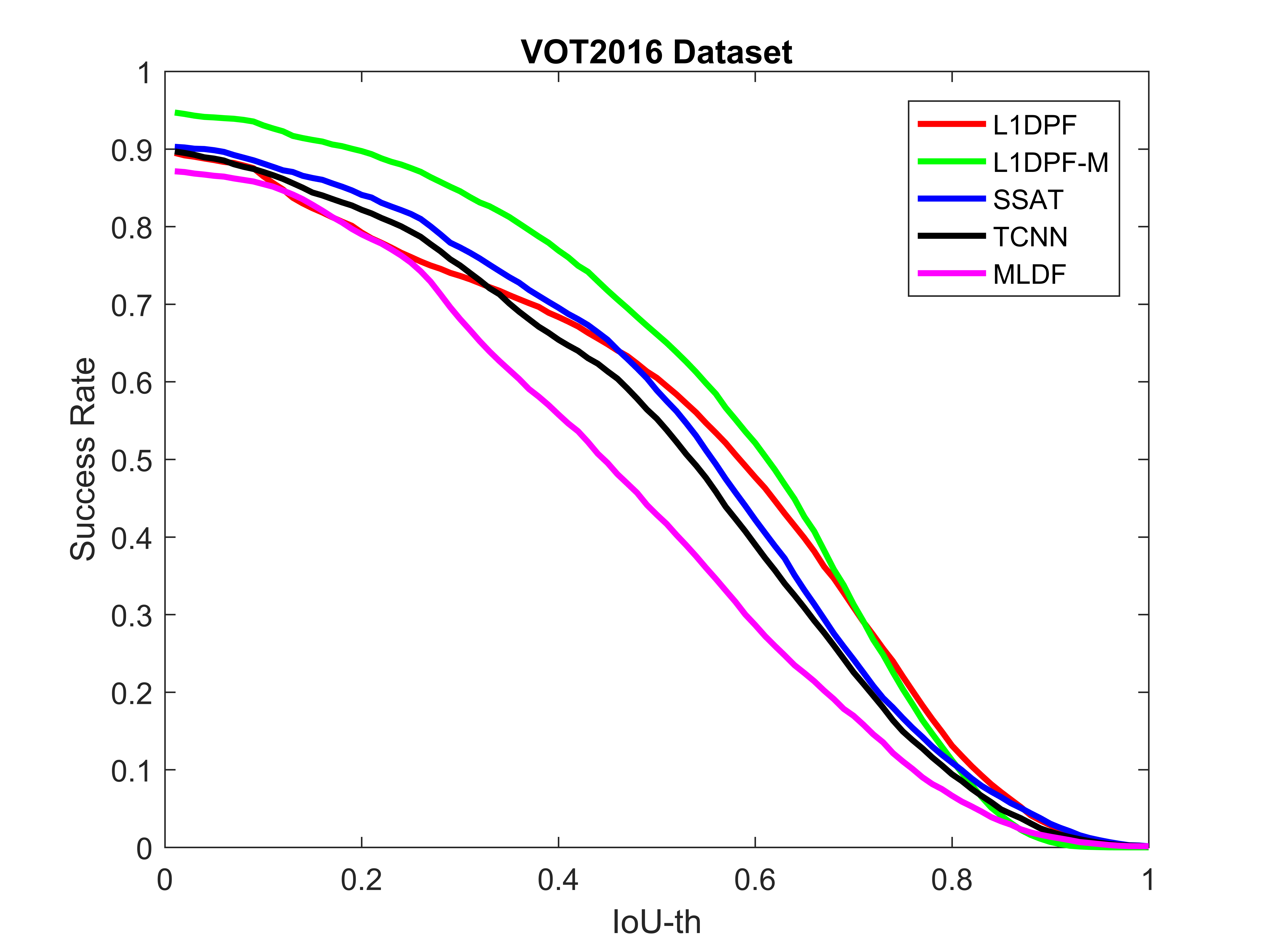}
    \caption{}
  \end{subfigure}
  \begin{subfigure}[ht]{2.4 in}
    \includegraphics[width=\linewidth]{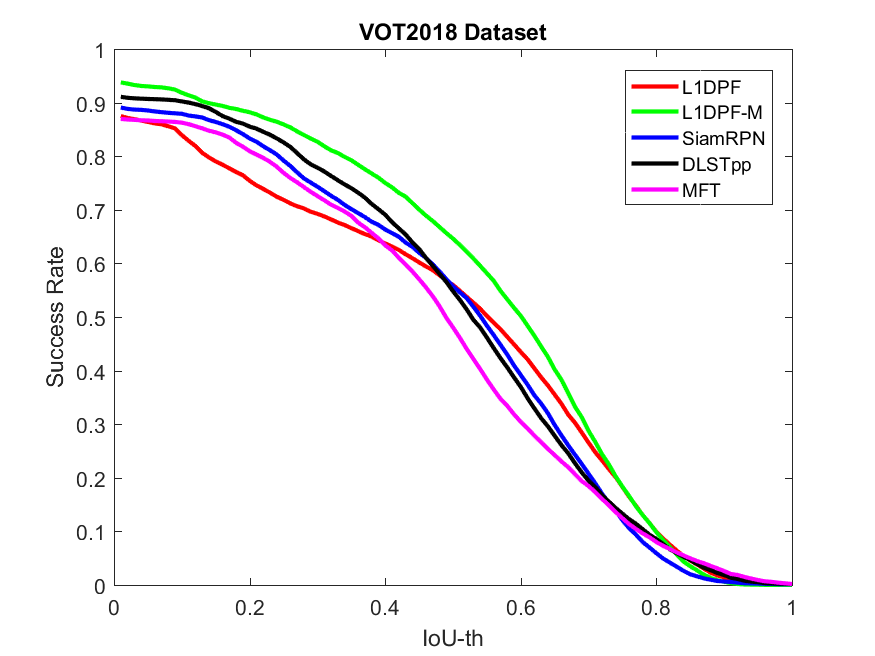}
   \caption{} 
  \end{subfigure}
 \caption{ Performance of L1DPF-M and L1DPF with respect to the top trackers of VOT2016 and VOT2018. Distribution of success rate versus IoU threshold. }
 \label{fig:votov}
\end{figure}

Success rates versus IoU-th are reported at Figure \ref{fig:votov}. Although the VOT benchmarking results are reported for reset and no-reset cases where the reset case allows re-initialization with the ground truth BB whenever the tracker fails \cite{VOT2016,VOT2018}, we tested our tracker for only no-reset case because it is much more realistic for applications.
Numerical results shown at Figure \ref{fig:votov} demonstrate that L1DPF-M provides the highest success rates at all IoU thresholds, because of the improved localization capability achieved by integration of the deep detector and the sparse object representation guided by a powerful dictionary update scheme. Success rates achieved by L1DPF are comparable to the-state-of-the-art trackers that illustrates the introduced integration model works properly. Performance is improved by the new state vector of L1DPF-M that enables us to efficiently track the deformed object BBs. In addition to the success rates we also report the accuracy and robustness of L1DPF-M compared to the top trackers of VOT2016 and 2018 at Table \ref{tab:accrob2016} and \ref{tab:accrob2018}, respectively. Scores for the top trackers are calculated by using the benchmarking results provided at \url{http://www.votchallenge.net/vot2016/results.html} and \url{http://www.votchallenge.net/vot2018/results.html}. Numerical results demonstrate that L1DPF-M provides the highest accuracy on both databases that indicates it achieves the highest mean IoU over the video sequences. Also L1DPF-M achieves the lowest robustness score on VOT2018 (Table \ref{tab:accrob2018}) and the second lowest score on VOT2016 (Table \ref{tab:accrob2016}) that illustrate that it does not frequently drifts off the target. This is because of the capability to track the deformed object BBs that enables tracking  longer video clips without dictionary update. High robustness scores reported for L1DPF confirm this conclusion, despite it takes second best place according to its accuracy scores.

\begin{table}[ht]
\caption{Accuracy, Robustness and Success Rate (SR) achieved by L1DPF-M/L1DPF on VOT2016 compared to the top trackers.}
\begin{center}
  \begin{tabular}{  l  c   c  c  c  c}
    \hline  
  	&L1DPF-M&L1DPF&TCNN&SSAT&MLDF\\ \hline
  Accuracy 	&0.60	&0.59&0.56&0.57&0.51\\ 
    Robustness	&0.10	&0.20 	&0.10 	&0.08 	&0.13 \\ 
    SR (IoU-th:0.5) & 0.66	& 0.61	&0.55  	& 0.59 	& 0.43 \\ 
 \hline

  \end{tabular}
  \label{tab:accrob2016}
\end{center}
\end {table}

\begin{table}[ht]
\caption{Accuracy, Robustness and Success Rate (SR) achieved by L1DPF-M/L1DPF on VOT2018 compared to the top trackers.}
\begin{center}
  \begin{tabular}{  l  c   c  c  c  c}
    \hline  
  	&L1DPF-M&L1DPF&SiamRPN&DLSTpp&MFT\\ \hline
  Accuracy 	&0.59	&0.57&0.56&0.56&0.56\\ 
    Robustness	&0.12	&0.24 	&0.17 	&0.14 	&0.19 \\ 
    SR (IoU-th:0.5) & 0.65	& 0.56	& 0.56 	& 0.55 	& 0.48 \\ 
 \hline

  \end{tabular}
  \label{tab:accrob2018}
\end{center}
\end {table}

\subsection{Attribute-based tracking accuracy} 
\begin{figure*}[ht!]
 \centering
  \begin{subfigure}[ht]{5.3in}
   \includegraphics[width=\linewidth]{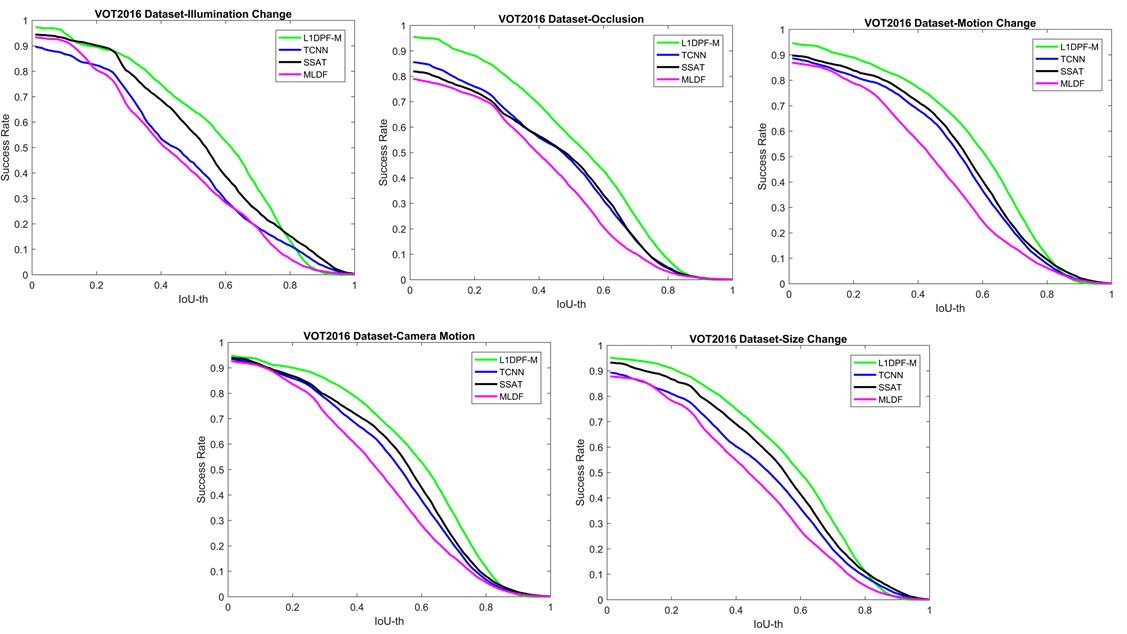}
    \caption{}
  \end{subfigure}
  \begin{subfigure}[ht]{5.3 in}
    \includegraphics[width=\linewidth]{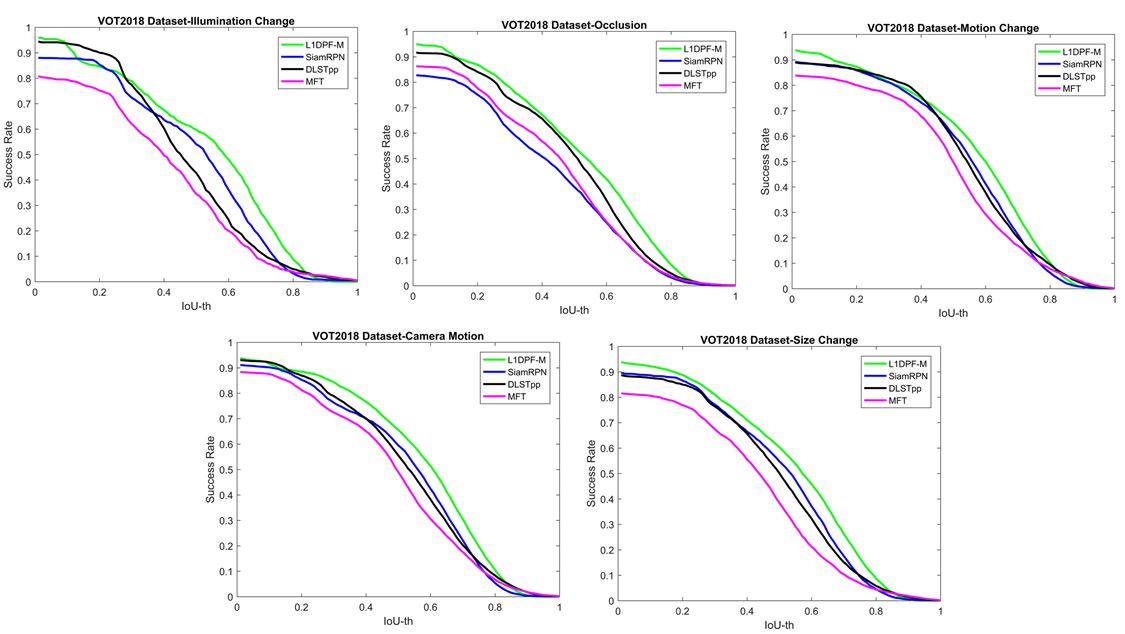}
   \caption{} 
  \end{subfigure}
 \caption{Attribute based evaluation on (a) VOT2016 data set, (b) VOT2018 data set. Success rate plots versus IoU threshold for different attributes: illumination change, occlusion,  motion change, camera motion and size change.}
 \label{fig:im4}
\end{figure*}

Since the tracking accuracy varies depending on the attribute type, we have also evaluated the attribute based performance. For a fair comparison, we made evaluation for occlusion (Occlusion), illumination change (Illumin.), motion change (Motion), size change (Size) and camera motion (Cam. M.), which are the five attributes used in VOT benchmarking. Table \ref{tab:att_Acc} reports the attribute based accuracy and robustness. Success rates achieved by L1DPF-M compared to the top trackers of VOT2016 and 2018 are illustrated in Figure \ref{fig:im4}.

It is clear that the most challenging attributes are camera motion and occlusion. Reported results indicate that L1DPF-M provides superior tracking accuracy for all attributes, but in particular for illumination change and size change where the improvement of success rate at IoU-th:0.5 is 9\% and 6\%  compared to SSAT and  5\% and 6\% compared to SiamRPN, respectively. Note that among the considered trackers SSAT is the one that estimates the object boundaries with deformed BBs, similar to us. In our opinion, this is why it achieves the highest rates compared to the others.    

\begin{table}[ht!]
\caption{Attribute based Accuracy(A)/Robustness(R)/Success Rate(SR) (IoU-th00.5) achieved by L1DPF-M compared to the top trackers of VOT 2016 and VOT 2018.}
\begin{center}
  \begin{tabular}{ c | l | c c c c| c c c c   }
    \hline  
&	& \multicolumn{4}{|c|}{\small{VOT 2016}} 	& \multicolumn{4}{c}{\small{VOT 2018}} \\ \hline
\small{Att}  &	&\footnotesize{L1DPF-M} &\footnotesize{TCNN}&\footnotesize{SSAT}&\footnotesize{MLDF}&\footnotesize{L1DPF-M}&\footnotesize{SiamRPN}&\footnotesize{DLSTpp}&\footnotesize{MFT}\\  \hline \hline
  \parbox[t]{1mm}{\multirow{3}{*}{\rotatebox[origin=c]{90}{\footnotesize{Illumin.}}}}   
  &  \small{A}	&0.557	&0.444 	&0.519	&0.430 &0.514	&0.463 	&0.451	&0.375	\\ 
&  \small{R}	&0.028	&0.103 	&0.056	&0.067 &0.041	&0.120 	&0.056	&0.192\\ 
&  \small{SR}	&0.647	&0.447 	&0.559	&0.403 &0.598	&0.54 	&0.427	&0.344\\ 
 \hline 
   \parbox[t]{1mm}{\multirow{3}{*}{\rotatebox[origin=c]{90}{\footnotesize{Occlusion}}}} 
  &  \small{A}	&0.503	&0.422 	&0.417	&0.37 &0.494	&0.389 	&0.465	&0.413	\\ 
&  \small{R}	&0.049	&0.144 	&0.181	&0.211 &0.054	&0.172 	&0.083	&0.137\\ 
&  \small{SR}	&0.558	&0.469 	&0.479	&0.361 &0.547	&0.389 	&0.514	&0.423\\ 
 \hline
 \parbox[t]{1mm}{\multirow{3}{*}{\rotatebox[origin=c]{90}{\footnotesize{Motion}}}} 
   &  \small{A}	&0.546	&0.478 	&0.5	&0.479 &0.53	&0.502 	&0.504	&0.455	\\ 
&  \small{R}	&0.057	&0.114 	&0.103	&0.132 &0.066	&0.107 	&0.110	&0.162\\ 
&  \small{SR}	&0.671	&0.559 	&0.593	&0.409 &0.653	&0.604 	&0.587	&0.498\\ 
 \hline
 \parbox[t]{1mm}{\multirow{3}{*}{\rotatebox[origin=c]{90}{\footnotesize{Cam. M.}}}} 
    &  \small{A}	&0.557	&0.491 	&0.512	&0.447 &0.545	&0.467 	&0.498	&0.456	\\ 
&  \small{R}	&0.058	&0.065 	&0.061	&0.074 &0.062	&0.088 	&0.069	&0.116\\ 
&  \small{SR}	&0.668	&0.559 	&0.611	&0.442 &0.656	&0.595 	&0.556	&0.49\\ 
 \hline
 \parbox[t]{1mm}{\multirow{3}{*}{\rotatebox[origin=c]{90}{\footnotesize{Size}}}} 
    &  \small{A}	&0.547	&0.412 	&0.51	&0.422 &0.57	&0.478 	&0.462	&0.397	\\ 
&  \small{R}	&0.048	&0.106 	&0.067	&0.123 &0.063	&0.103 	&0.113	&0.184\\ 
&  \small{SR}	&0.638	&0.504 	&0.573	&0.424 &0.604	&0.543 	&0.502	&0.385\\ 
 \hline 
  \end{tabular}
  \label{tab:att_Acc}
  \end{center}
\end {table}

\begin{figure*}[ht!]
 \centering
  \begin{subfigure}[ht]{4.6 in}
   \includegraphics[width=\linewidth]{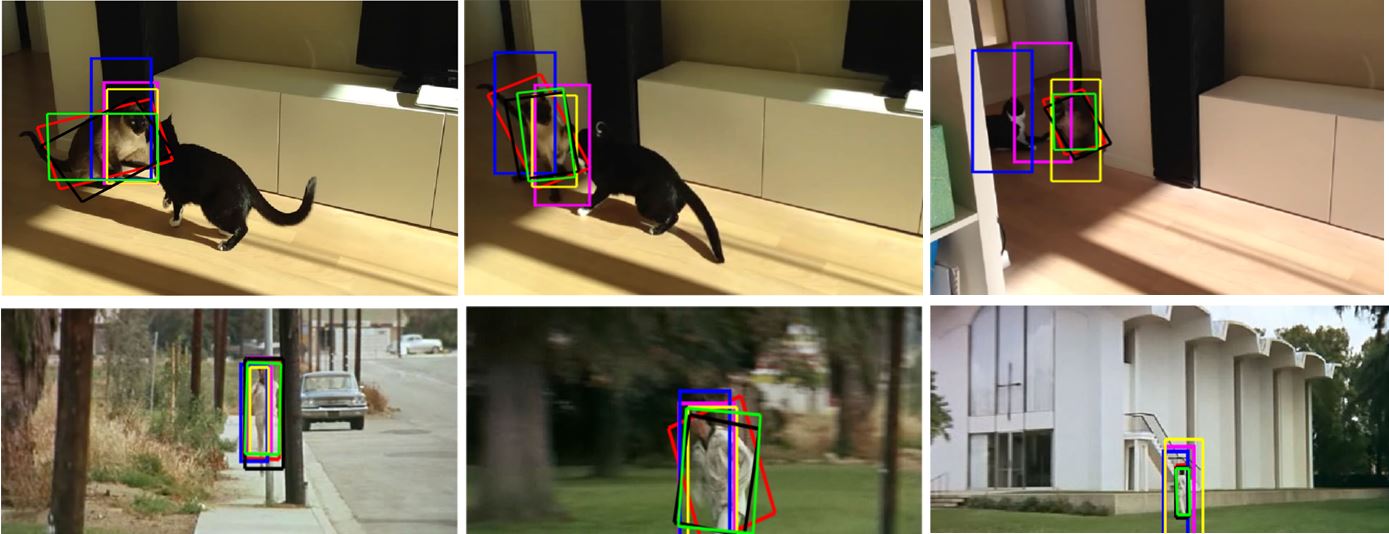}
    \caption{}
  \end{subfigure}
  \begin{subfigure}[ht]{4.6 in}
    \includegraphics[width=\linewidth]{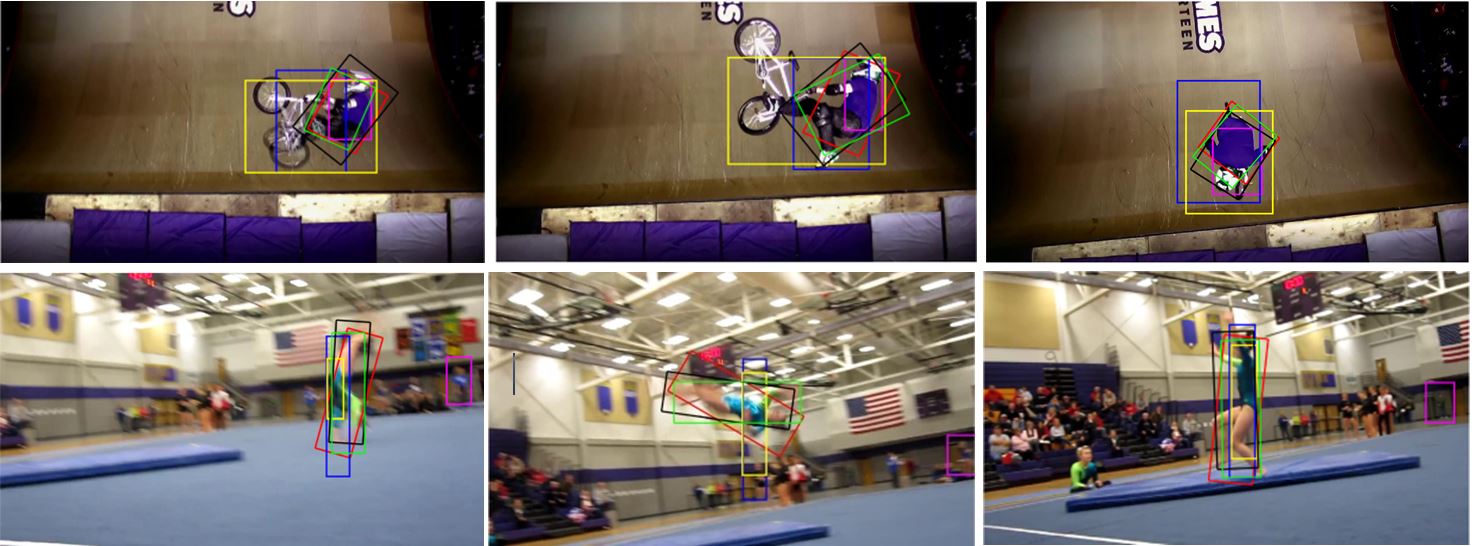}
   \caption{} 
  \end{subfigure}
 \caption{Object BBs tracked by L1DPF-M(green), L1DPF (red), GT(black) and top trackers of (a) VOT2016, TCNN(magenta), SSAT(yellow), MLDF(blue). fernando, frame no: 156,179,233 graduate, frame no: 262,544,611 (b) VOT2018, DLSTpp(yellow), SiamRPN (magenta), MFT(blue). bmx, frame no: 19,27,65, gymnastics2, frame no: 188,214,226}
 \label{fig:imgg}
\end{figure*}

In order to visualize differences from the existing trackers, we report the tracked object BBs on a number of video frames. Four challenging video sequences, namely, Fernando, Graduate, Bmx and Gymnastics3 are chosen from VOT2016/2018 benchmark data sets. Figure \ref{fig:imgg}(a) illustrates the object BBs tracked by the proposed tracker L1DPF-M compared to TCNN, SSAT, and MLDF which are the top trackers of VOT2016.  BBs tracked by L1DPF are also visualised for comparison. In Fernando sequence, all trackers catch the target object in frame 156 and 179, but L1DPF-M provides higher localization accuracy with its deformed BB. Before frame 233, L1DPF and L1DPF-M  update the dictionary at frame 227 and 229 respectively, because of illumination change and occlusion and by the help of superior localization capability of deep detector, both trackers keep tracking with high localization until frame 239 and 243. In Graduate sequence, proposed tracker samples more accurate particles and more robust to self occlusion at frame 262 as a result of the introduced observation model. Also appearance and scale changes encountered at frame 544 and 611 are well handled because of the introduced target model update scheme. At frame 544, it is clear that because of the proposed state vector, estimation of L1DPF-M is more accurate  than the others.

Figure \ref{fig:imgg}(b) illustrates improvement achieved by L1DPF-M on Gymnastics2 and Bmx sequences compared to DLSTpp, SiamRPN and MFT where the object tracking becomes harder because ıf the abrupt appearance changes and rotations. In video frames of Bmx sequence, all trackers keep tracking but L1DPF-M achieves the highest IoU with the GT as a result of the proposed state vector that enables to estimate object boundaries with a deformed BB.  
In video frames of Gymnastics2 sequence, other trackers fail to localize the person at frame 214, whereas L1DPF-M accurately adapts the orientation of the tracked BB according to the object motion. This is because the proposed state vector enables us to correctly estimate the object motion, in this case rotation.

\section{Conclusions}
\label{conc}

We introduce a tracking by detection method that takes the advantages of Bayesian filtering as well as the deep learning. Attribute based performance of L1DPF-M demonstrates that integration of the deep detector into  the tracking model significantly improves robustness to scale changes. Specifically instance segmentation achieved by the deep detector provides more accurate object boundaries. Robustness to scale changes increases with the deformed object BB tracking capability of PF provided by the affine motion representative state vector. Moreover particle filtering notably increases the tracking performance under illumination changes. This is mainly because of the simple motion model of PF that enables efficient modelling of the temporal video content and yields an uninterrupted detection. In order to improve robustness to appearance changes, we also propose a dictionary update scheme that effectively monitors the sufficiency of dictionary. This allows L1DPF-M to track the object without drifting. Experimental results on challenging video sequences show that the proposed tracker, L1DPF-M, achieves the highest tracking accuracy compared to the state-of-the-art trackers, while robustness, that indicates the number of failures, is the lowest.

\bibliography{mybible}

\end{document}